\documentclass{article}

     \usepackage[preprint,nonatbib]{neurips_2020}

\usepackage[numbers]{natbib}
\usepackage[utf8]{inputenc} 
\usepackage[T1]{fontenc}    
\usepackage{hyperref}       
\usepackage{url}            
\usepackage{booktabs}       
\usepackage{amsfonts}       

\usepackage{microtype}      
\usepackage{graphicx}
\usepackage{nicefrac}       
\usepackage{amsthm}
\usepackage{amsmath}
\usepackage{amsfonts}
\usepackage{bbm}
\usepackage{amssymb}
\usepackage{wrapfig}
\usepackage{xcolor}

\usepackage{float}        
\usepackage{subcaption}     
\usepackage{graphicx}      
\usepackage{dcolumn}
\usepackage{bm}
\usepackage{color}

\usepackage{todonotes}

\usepackage{enumitem}
\usepackage{algorithm}
\usepackage{algorithmic}
\usepackage{footnote}
\DeclareMathVersion{normal2}

\usepackage{amsmath,amsfonts,bm}

\def\eqref#1{equation~\ref{#1}}

\def\1{\bm{1}}

\def\vg{{\bm{g}}}

\def\vw{{\bm{w}}}
\def\vx{{\bm{x}}}
\def\vy{{\bm{y}}}
\def\vz{{\bm{z}}}

\def\mA{{\bm{A}}}
\def\mB{{\bm{B}}}

\def\mG{{\bm{G}}}
\def\mH{{\bm{H}}}

\def\mM{{\bm{M}}}

\def\mP{{\bm{P}}}

\def\mR{{\bm{R}}}

\def\mW{{\bm{W}}}
\def\mX{{\bm{X}}}
\def\mY{{\bm{Y}}}

\def\mr{{\bm{\mathcal{R}}}}
\def\mtx{{\bm{\tilde{X}}}}

\DeclareMathAlphabet{\mathsfit}{\encodingdefault}{\sfdefault}{m}{sl}
\SetMathAlphabet{\mathsfit}{bold}{\encodingdefault}{\sfdefault}{bx}{n}

\newtheorem{definition}{Definition}[section]
\newtheorem{theorem}{Theorem}

\theoremstyle{remark}

{
	\theoremstyle{plain}
	
}

\title{Beyond Random Matrix Theory for Deep Networks}

\author{%
  Diego Granziol\thanks{\href{http://www.granziol.me}{Personal Homepage}} \\
Machine Learning Research Group\\
Oxford\\
United Kingdom \\
\texttt{diego@robots.ox.ac.uk} \\
}

\begin{document}

\maketitle
\begin{abstract}
We investigate whether the Wigner semi-circle and Marcenko-Pastur distributions, often used for deep neural network theoretical analysis, match empirically observed spectral densities. We find that even allowing for outliers, the observed spectral shapes strongly deviate from such theoretical predictions. This raises major questions about the usefulness of these models in deep learning. We further show that theoretical results, such as the layered nature of critical points, are strongly dependent on the use of the exact form of these limiting spectral densities. We consider two new classes of matrix ensembles; random Wigner/Wishart ensemble products and percolated Wigner/Wishart ensembles, both of which better match observed spectra. They also give large discrete spectral peaks at the origin,
providing a theoretical explanation for the observation that various optima can be connected by one dimensional of low loss values. We further show that, in the case of a random matrix product, the weight of the discrete spectral component at $0$ depends on the ratio of the dimensions of the weight matrices.
\end{abstract}

\section{Introduction}
\label{introduction}

Deep neural networks have managed to achieve state of the art performance for tasks in computer vision, natural language processing, speech recognition, reinforcement learning and drug discovery \citep{mayr2016deeptox,lecun2015deep}. However 
given the extremely high dimensional and non convex loss surfaces of neural networks, it is still not understood why gradient based methods \cite{Bottou2012}, with different random initializations and hyper-parameters do not seem to get stuck in poor quality local minima or saddle points? 
Some researchers have considered making several strong simplifying assumptions, such as independence of the neural network inputs and weights \citep{choromanska2015loss,choromanska2015open,pennington2017geometry}. under these assumptions, critical points (where the gradient is zero) of the loss, are described by certain classes of well studied stochastic matrices, such as the Gaussian Orthogonal or Wishart Ensembles \citep{bun2017cleaning}. The spectra of these matrices, in the limit of infinite dimension, are known and this subject of enquiry is known as \textit{random matrix theory}. Hence under the guise of these assumptions, we can make quantitative predictions about the nature of the critical points and the general geometry of the loss landscape, as described by the Hessian of the loss, which we introduce formally in section \ref{subsec:hessianprimer}. 

\citet{choromanska2015loss} show for i.i.d Gaussian inputs and network path independence, that a multi-layer neural networks loss is equivalent to that of a spin-glass model. Its Hessian spectrum is thus given by the Gaussian Orthogonal Ensemble \citep{auffinger2013random}, a random matrix where each element is sampled independently from a Gaussian with common variance. Local minima are thus located within a narrow band, lower bounded by the global minimum. the practical implication is that for a sufficient number of hidden layers (more than 2) all local minima are \textit{close} in loss to the global. \citet{pennington2017geometry} use the Gauss Newton decomposition of a squared loss Hessian, assuming independence and normality of both the data and weights, along with free addition of the resulting Wigner/Wishart ensembles, they derive a functional form for the critical index (fraction of the eigenvalues are negative) as a function of the loss. They show that below a certain critical energy threshold \textit{all critical points are minima}. In \citet{ba2020generalization} they assume Gaussian inputs, i.i.d Gaussian weights and a linear teacher model and use random matrix theory to derive the generalisation properties of two layer neural networks, such as the population risk\footnote{Loss under the expectation of the data generating distribution.} of the regularised least squares problem and to derive the \textit{double descent phenomenon}. Increasing the network size initially leads to over-fitting, but beyond a critical point, further increasing the network size decreases the test error and to a lower level than the optimal small network. 

This paper extensively empirically evaluates whether the random matrix ensembles employed in theory match with real neural network spectra at the end of training. We show major discrepancies between what is predicted/assumed theoretically and what is observed in practice. This raises serious questions about the validity and usefulness of these models for explaining observed neural network phenomena. We expressly show that if the assumption of identical distribution of weights or data is dropped, or the spectrum is allowed to have certain large correlations due to outliers,that the claims of layered loss structure in \citet{choromanska2015loss} no longer hold. Outliers are extensively empirically observed \citep{papyan2018full,papyan2019measurements,ghorbani2019investigation,sagun2016eigenvalues,sagun2017empirical}. We investigate promising alternatives in the random matrix theory literature, in particular percolated and product of matrix spectra, which well match observed neural network spectra.


\subsection{Motivation}
\label{sec:annaisntonthemoney}

The ability to ablate the unrealistic assumptions of both weight and input independence is considered a major open research problem \citep{choromanska2015open} for theoretical deep learning, with some success in the deep linear case \citep{kawaguchi2016deep}, although this model is still argued to be unrealistic \citep{nguyen2017loss}. Given that there exists no general theory of dependent matrix ensembles, achieving the same theoretical results under a less restrictive framework, is potentially intractable. Instead in this paper, we empirically evaluate concrete theoretical predictions, specifically whether the neural network Hessian spectrum is similar to proposed ensembles such as the GOE or Wishart. If these spectra are well matched, intuitively the dependence structure does not invalidate the final result. This idea, that we can experimentally validate a theory even with what we consider to be unrealistic assumptions, was central to the foundation of random matrix theory, developed by \citet{wigner1958distribution} in the context of nuclear scattering. This line of thinking was drawn from statistical mechanics \citep{landau_statphys}, where sharp quantitative predictions were made by focusing on a statistical description of the problem, enforcing conservation laws as constraints. Wigner hence replaced the Hamiltonian with a random matrix, which respected Hermicity and time-reversal invariance.\footnote{Which means the Hamiltonian must be real.} The simplest matrix satisfying such conditions is one with elements drawn independently from the Gaussian distribution known as the \textit{Gaussian Orthogonal Ensemble (GOE)}. He was able to relate the energy level spacing in the nuclei to the spacings between the GOE eigenvalues, achieving excellent experimental agreement \citep{moldauer1961theory}. The key insight was that particle interactions, although prevalent and strong, are not relevant for the overall system description. This idea allowed statistical mechanics to avoid solving multiple body Newtons equations and in nuclear scattering, the many body Schr\"{o}dinger equation, both of which are intractable.

\begin{figure}[h]
    \centering
    \begin{subfigure}{0.32\linewidth}
        \includegraphics[width=1\linewidth,trim={0 0 0 0},clip]{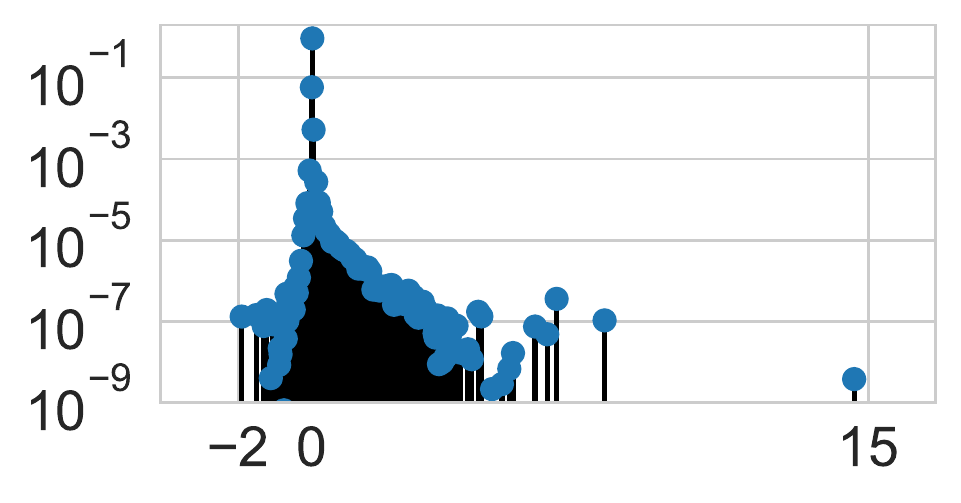}
        \caption{$\mH(\vw)$ VGG$16$BN epoch $300$}
        \label{subfig:vgg16bnhess300}
    \end{subfigure}
    \begin{subfigure}{0.32\linewidth}
        \includegraphics[width=1\linewidth,trim={0 0 0 0},clip]{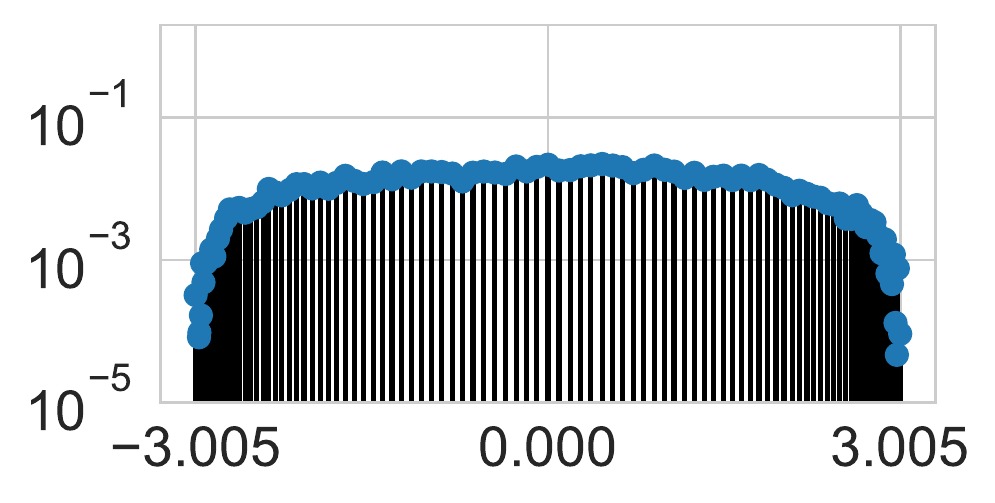}
        \caption{Wigner semi-circle}
        \label{subfig:wignerstemintro}
    \end{subfigure}
    \begin{subfigure}{0.32\linewidth}
        \includegraphics[width=1\linewidth,trim={0 0 0 0},clip]{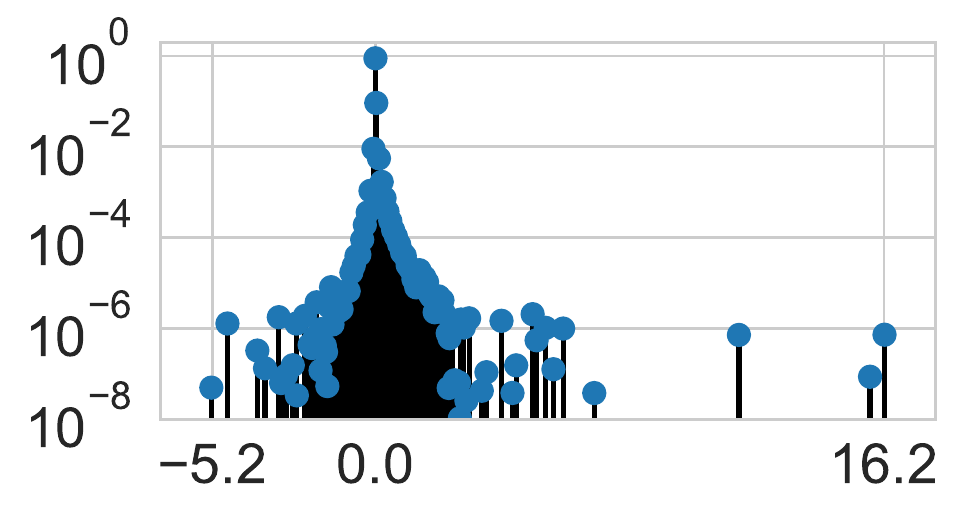}
        \caption{$\mH(\vw)$ PResNet-$110$ epoch $300$}
        \label{subfig:preresnet110300}
    \end{subfigure}
    \caption{Neural network Hessian spectra at the end of training for the VGG-$16$BN and PreResNet-$110$ on the CIFAR-$100$ dataset and an example of  the semi-circle law}\label{fig:goeisabadfit}
\end{figure}

Applying the same logic to neural networks, we would similarly expect the Hessian spectrum, if the spin glass model from \citet{choromanska2015loss,auffinger2013random} is relevant, to resemble that of the GOE, which is given by semi-circle law, shown in Figure \ref{subfig:wignerstemintro}. However for real world neural networks on a typical dataset, such as the VGG-$16$ with batch-normalisation \citep{simonyan2014very,ioffe2015batch} and the PreResNet-$110$ \citep{he2016deep} at the end of training on the CIFAR-$100$ dataset, we see from Figure \ref{fig:goeisabadfit} significant deviations between the semi-circle law as predicted by \citep{choromanska2015loss}, notably the seeming rank degeneracy at the origin, the asymmetry of the bulk and the outliers.
Given the aberration between theoretical predictions and empirical observations, this raises major questions about the usefulness or the empirical validity of using these models in the context of deep learning. 
Further investigation of the deviation from and the importance of convergence to the assumed limiting spectral distribution in previous theoretical work, is required. As we go on to show in this paper, even several subtle changes, for which we would still expect the Hessian to converge to the Wigner semi-circle density, would change the strength of the theoretical results. Adding outliers, through the use a spiked model \citet{benaych2011eigenvalues}, can be shown to invalidate certain theoretical results. This work challenges the notion that these theoretical results hold generally, even if the assumptions used are quite strict.

This papers main contributions are 
\begin{itemize}
    \item A specific demonstratation that altering the limiting spectral densities by adding outliers or changing the nature of the ensemble (such as removing the identical variances of the Hessian elements) directly invalidates theoretical loss surface claims made in \citet{choromanska2015loss}
    \item An empirical investigation into the similarity between Hessian, Generalised Gauss Newton and Residual matrices and their purported random matrix counterparts, the Gaussian Orthogonal and Wishart ensembles \citep{choromanska2015loss,pennington2017geometry,sagun2017empirical,pedregosa2020average}
    \item An investigation into both the random sparse graph model and random product model for the Hessian, both of which show a promising similarity to real neural network Hessia. Furthermore, we show that these models further predict strange properties observed in neural networks such as the inter-connectivity of low loss areas in the weight-space \citet{garipov2018loss,izmailov2018averaging} and that this depends on the dimensionality of the matrix products.
\end{itemize}


\section{Background Theory}
\label{subsec:rmt}
\label{sec:backgroundtheory}
The eigenvalue density of a specific realisation of a random matrix can be characterised by its empirical spectral distribution, or equivalently its Stieltjes transform
\begin{equation}
    \rho_{M}^{P}(x) = \frac{1}{P}\sum_{i=1}^{P}\delta(x-\nu_{i}), \thinspace    \mathcal{S}_{N}(z) = \frac{1}{N}\text{Tr}[(zI_{N}-H)^{-1}] \xrightarrow{N \rightarrow \infty}   \int \frac{\rho(u)}{z-u}du \thinspace z \in \mathbb{C}
\end{equation}


For large random matrices one expects the empirical spectral density to converge $\rho_{M}^{P}\rightarrow\rho_{M}$ as $P \rightarrow \infty$. A crucial feature of random matrix theory is the self-averaging property (known as ergodicity or concentration property) of the limiting spectral density \citep{bun2017cleaning}. As the matrix dimension $P \rightarrow \infty$ a single realization spans the whole eigenvalue density function. The \textbf{Wigner semi-circle} and \textbf{Marcenko-Pastur} are the limiting spectral densities for which a large class of random matrices converge \citep{yin1986limiting,silverstein1995strong} and hence have a privileged role in RMT related analysis. \paragraph{The Wigner Semi-Circle Law} arises from matrices, for which the elements are drawn independently from zero mean unit variance distributions. In the special case of the normal distribution $\mH_{i,j} = \mathcal{N}(0,\sigma^{2})$, this is known as the \textbf{Gaussian Orthogonal Ensemble} which gives the limiting spectral density 
\begin{equation}
\label{eq:babywigner}
     p(\lambda) = \frac{1}{2\pi}\sqrt{4\sigma^{2}-\lambda^{2}}\mathbf{1}_{|\lambda|\leq 2\sigma}
\end{equation}
The Wigner matrix, 
has been used to analyse the residual component $\mr$ (which we define in the next section) \citep{pennington2017geometry} and \citet{choromanska2015loss} use the GOE as a model of the Hessian.
\paragraph{the Marcenko-Pastur law} arises from the product of random matrices $\mH = \frac{1}{N}\mX\mX^{T}$, where $\mX \in \mr^{P\times N}$ and  $\mX_{i,j}$ are zero mean with variance $\sigma^{2}$, defining $q=P/N$, the limiting spectral density is given as
\begin{equation}
\label{eq:babymp}
  \begin{cases}
    (1-\frac{1}{q})\mathbbm{1}_{0\in A}+\nu_{1/q}(A) ,& \text{if } q> 1 \\
    \nu_{q}(A) ,& \text{if } 0\leq q \leq 1 
    \end{cases} 
    , d\nu_{q} =\frac{\sqrt{(\lambda-\lambda_{+})(\lambda-\lambda_{-})}}{ 2\pi\lambda q \sigma^{2}}
    , \lambda_{\pm} = \sigma^{2}(1\pm \sqrt{q})^{2}
\end{equation}
The Marcenko-Pastur, 
due to its low rank positive definite nature, has been widely adopted to analyse the generalised Gauss-Newton \citep{sagun2016eigenvalues,sagun2017empirical,pennington2017geometry,pedregosa2020average}. 


\label{subsec:hessianprimer}
\subsection{Neural network Hessians}
For an input/target pair $[\vx \in \mathbb{R}^{d_{x}},\vy \in \mathbb{R}^{d_{y}}]$ and a given model output $h(\cdot;\cdot):\mathbb{R}^{d_{x}}\times \mathbb{R}^{P} \rightarrow \mathbb{R}^{d_{y}}$, parameterised by the weight vector $\vw$, i.e., $\mathcal{H}:= \{h(\cdot;\vw):\vw \in \mathbb{R}^{P} \}$, with a given loss function $\ell(h(\vx;\vw), \vy): \mathbb{R}^{d_{y}} \times \mathbb{R}^{d_{y}} \rightarrow \mathbb{R}$. 
The empirical risk, denoted the loss in deep learning, is given by $R_{emp}(\vw) = \frac{1}{N}\sum_{i=1}^{N}\ell(h(\vx_{i};\vw),\vy_{i})$
For a dataset of size $N$, with gradients $\vg_{emp}(\vw)$ and Hessians $\mH_{emp}(\vw)$ thereof. The Hessian describes the curvature at that point in weight space $\vw$ and hence the risk surface can be studied through the Hessian. 
For a quadratic loss function $\frac{1}{N}\sum_{i}^{N}\ell(h(\vx_{i};\vw),\vy_{i}) = \frac{1}{2N}\sum_{i}^{N}||h(\vx_{i};\vw)-\vy_{i}||^{2} = \frac{1}{2N}\sum_{i}^{N}r_{i}^{2}$, the elements of the Hessian are given by
\begin{equation}
    \mH_{j,k} = \frac{1}{N}\sum_{i}^{N}\bigg(\frac{\partial r_{i}}{\partial \vw_{j}}\frac{\partial r_{i}}{\partial \vw_{k}}+r_{i}\frac{\partial^{2} r_{i}}{\partial \vw_{j}\partial \vw_{k}}\bigg) = \mG_{j,k}+\mr_{j,k}
\end{equation}
Where the first term on the right hand side is known as the Gauss-Newton matrix \footnote{the Gauss Newton method proceeds by approximating $\mH$ with $\mG$} and we denote the \textit{Residual matrix} $\mr$. In this case where the model perfectly interpolates the data, the residuals $r_{i} = 0 \forall i$ and $\mG = \mH$. As $\mG$ is the sum of a dyadic product, it is trivially positive semi-definite and has a rank of at most $N$.  In deep learning classification tasks the loss function $\ell(h(\vx;\vw), \vy)$ is the cross entropy loss, which is typically matched with a softmax activation, such that the per class output can be interpreted as a probability.
\begin{equation}
\label{eq:crossentropy}
    \frac{1}{N}\sum_{i}^{N}\ell(h(\vx_{i};\vw),\vy_{i}) = \frac{1}{N}\sum_{i}^{N}\sum_{c=1}^{d_{y}}\mathbbm{1}_{i,c}\log h(\vx_{i};\vw)_{c}, \thinspace \& \thinspace h(\vx_{i};\vw)_{c} =  \frac{\exp(\vz(\vx_{i};\vw)_{c})}{\sum_{k=1}^{d_{y}}\exp(\vz(\vx_{i};\vw)_{k})}
\end{equation}
Where $d_{y}$ is the number of classes and $\mathbbm{1}_{i,c}$ is the indicator function which takes the value of $1$ for the correct class and $0$ for the incorrect class, $\vz(\vx_{i};\vw)$ is the softmax input. By writing the empirical risk in terms of the activation $\sigma$ at the output of the final layer $f(\vx)$. The Hessian may be expressed using the chain rule as \begin{equation}
\label{eq:fullHessian}
\begin{aligned}
 & \mH(\vw)_{jk} =
\frac{1}{N}\sum_{n=1}^{N}\bigg(  \sum_{c=0}^{d_{y}} \sum_{l=0}^{d_{y}} \frac{\partial^{2} \sigma(f(\vx))}{\partial f_{l}(\vx)\partial f_{c}(\vx)}\frac{\partial f_{l}(\vx)}{\partial x_{j}}\frac{\partial f_{c}(\vx)}{\partial x_{c}} 
+ \sum_{c=0}^{d_{y}}\frac{\partial \sigma(f(\vx))}{\partial x_{j}} \frac{\partial^{2} f_{c}(\vx)}{\partial x_{j}\partial x_{k}} \bigg)\\
\end{aligned}
\end{equation}
The first term on the LHS of \eqref{eq:fullHessian} is known as the generalised Gauss Newton (GGN) matrix \citep{schraudolph2002fast,kunstner2019limitations,martens2010deep} which is often referred to as the Gauss Newton in the deep learning literature. The rank of a product is the minimum rank of its products so the GGN is upper bounded by the $N\times d_{y}$. The Residual matrix $\mr$, has not been systematically investigated, except in \citep{papyan2019measurements}, in which it was shown not to contribute to the outliers.

\section{The Problem with Random Matrix Theory and DNN Hessian Spectrums}
\label{sec:rmtissues}
This section investigates the validity of several random matrix theoretic assumptions employed in theoretical Deep Learning. Specifically, section \ref{subsec:GOEismagical} investigates the effects of small perturbations to the Gaussian Orthogonal Ensemble, such as adding outliers to the spectrum, or using the more general Wigner ensemble. Section \ref{subsec:ggn} investigates whether the generalised Gauss-Newton spectrum resembles the Marcenko-Pastur and whether the Residual matrix spectrum. resembles the Wigner semi-circle matrices as assumed in \citet{pennington2017geometry}.

\subsection{The Magic of the Gaussian Orthogonal Ensemble}
\label{subsec:GOEismagical}
As discussed in Section \ref{subsec:rmt}, the GOE is a very specific subset of Wigner matrices. Its spherical symmetry of the matrix allowing for elementary proofs of otherwise complicated results \cite{bun2016rotational,bun2017cleaning}. Although we do not investigate the claim in depth here, we note that spherical symmetry is relied upon in \citet{ba2020generalization}.  Let us consider the following claims made in \citet{choromanska2015loss} and the evolving mathematics under perturbations to the GOE.

\begin{description}[font=$\bullet$~\normalfont\scshape\color{red!50!black}]
\item [Claim 1]  
\textit{Above a certain loss threshold any critical point is a high-index saddle point w.o.p\footnote{with overwhelming probability}}
\item [Claim 2] 
\textit{The loss surface possesses a layered structure w.o.p}
\end{description}
Claim 1, relies on Theorem 2.14 \citep{auffinger2013random}, which requires the large deviation principle for the GOE \citep{arous1997large}. Specifically, for some $C_{\epsilon}>0$ 
\begin{equation}
\label{eq:wearenotoverwhelmed}
    \mathbb{P}\bigg[ \lambda_{k} \geq -\sqrt{2} + \epsilon\bigg] \leq e^{-C_{\epsilon}P^{2}}
\end{equation}
Where in this formulation, the variance is given by $1/2P$ and hence the support of the semi-circle is $[-\sqrt{2},\sqrt{2}]$. The relevant proof in \citep{arous1997large} rests heavily on the rotational invariance of the i.i.d Gaussian entries assumption to achieve the exponential concentration in equation \ref{eq:wearenotoverwhelmed}. For general Wigner matrices, not restricted to the i.i.d Gaussian case, which would result from having non Gaussian distributed data or weights, or a mix of independent Gaussians of different variances, the convergence is known to be $\mathcal{O}(N^{-1/4})$ \cite{bai2003convergence} and hence the term \textit{with overwhelming probability} no longer holds. In this case we would still expect the LSD to converge to the semi circle law.
Now let us ignore Figure \ref{fig:goeisabadfit} and suppose for ease of exposition that the bulk was in agreement with the semi-circle law, with the exception of a few outliers. How is the concentration in \eqref{eq:wearenotoverwhelmed} explicitly altered by this perturbation? The largest $k$ eigenvalues for small $k$ is given by the outliers, which for the perturbed Gaussian Orthogonal Ensemble, is proportional to the number of parameters $P$ \citep{granziol2020towards,benaych2011eigenvalues} as opposed to the bulk which is proportional to $\sqrt{P}$. We derive this result in full in Appendix \ref{sec:beastderivation}.
\begin{equation}
	\label{eq:spectralbroadeningwignernoise}
	\lambda'_{1} = \left\{\begin{array}{lr}
		\lambda_{1} + \sqrt{\frac{P}{2N}}\frac{\sigma}{\lambda_{1}}, & \text{if } \lambda_{1} > \sqrt{2}\\
		\sqrt{2}, & \text{otherwise } \\
	\end{array}\right\} \thinspace , \thinspace \lambda'_{P} = \left\{\begin{array}{lr}
			\lambda_{P} + \sqrt{\frac{P}{2N}}\frac{\sigma}{\lambda_{P}}, & \text{if } \lambda_{P} < -\sqrt{2}\\
		- \sqrt{2}\sigma, & \text{otherwise } \\
	\end{array}\right\}.
\end{equation}
and hence the edge of the spectrum grows with $P$, so the large deviation principle cannot be used to bound the spectrum with high probability, this time not just weakening claim $1$, but invalidating it entirely. Claim 2, which states that for large networks, critical values above the global minimum follow a banded structure where between $L_{0},L_{1}$ there are only local minima and then between $L_{1},L_{2}$ there are only local minima and critical points of order 1, etc relies on Theorems 2.15, which further necessitates theorems 2.5 and 2.1 from \citep{auffinger2013random}. Specifically Theorem 2.5 relies on the $k+1$'th smallest eigenvalue satisfying the good rate function of the GOE \citep{arous1997large}. We can repeat the proceeding arguments for the $k$-largest eigenvalues being outliers, but for negative outliers as per equation \ref{eq:spectralbroadeningwignernoise}. Note from \ref{fig:goeisabadfit}, that in real world neural networks, we observe negative eigenvalues seperated from spectral bulk, hence their theoretical inclusion is necessitated. The explicit dependence of these theorems on the GOE and specifically the extremal eigenvalues of the GOE, is given in the following theorems
\begin{theorem}
(Theorem 2.1 in \citet{auffinger2013random})The following identity holds for all $N, p \geq 2, k \in \{0,...,N-1\}$, and for all Borel sets $B \in \mathbb{R}$
\begin{equation}
    \mathbb{E}[\text{Crt}_{N,k}(B)] = 2 \frac{2}{p}(p-1)^{N/2}\mathbb{E}^{N}_{GOE}\bigg[e^{-N\frac{p-2}{2}(\lambda_{k}^{N})^{2}}\mathbbm{1}\bigg\{\lambda_{k}^{N}\in \sqrt{\frac{p}{2(p-1)}B}\bigg\}\bigg]
\end{equation}
\end{theorem}
and for theorem 2.5
\begin{equation}
\lim_{N \rightarrow \infty} \frac{\log \mathbb{E}[\mathrm{Crt}_{N,k}(u)]}{N} = \frac{1}{2}\log (p-1) + \lim_{N \rightarrow \infty} \frac{1}{N}\log \mathbb{E}^{N}_{GOE}[e^{-N(p-2)\lambda_{k}^{2}/2p}\mathbbm{1}_{\lambda_{k}\leq t}]
\end{equation}
where $\lambda_{k}$ is the $k$'th smallest eigenvalues of $\mH$ and $\mathrm{Crt}_{N,k}$ denotes the number of critical points of order $k$ when the matrix of size $N$.
\subsection{Generalised Gauss Newton and Residual matrices}
\label{subsec:ggn}

\begin{figure}[h!]
    \centering
    \begin{subfigure}{0.24\linewidth}
        \includegraphics[width=1\linewidth,trim={0 0 0 0},clip]{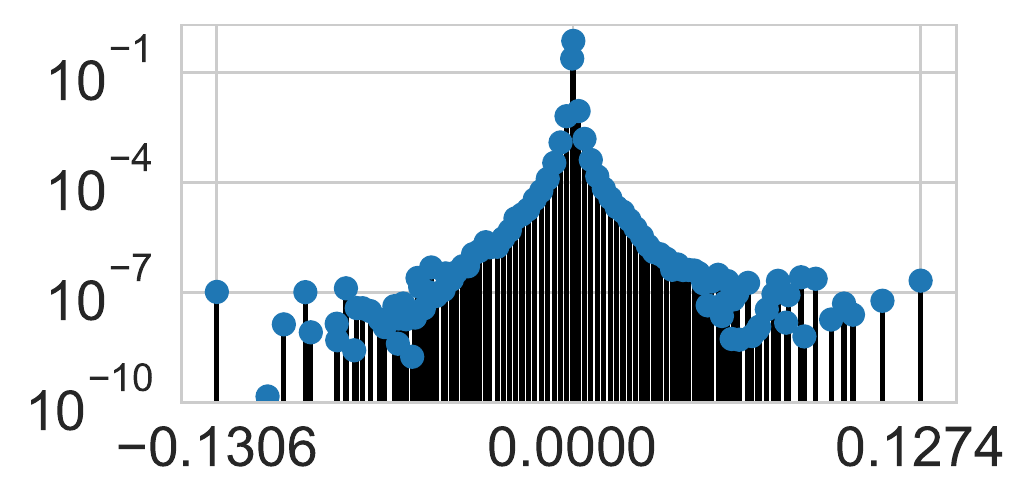}
        \caption{$\mr$ VGG-$16$ Epoch $300$}
        \label{subfig:vgg16res300}
    \end{subfigure}
    \begin{subfigure}{0.24\linewidth}
        \includegraphics[width=1\linewidth,trim={0 0 0 0},clip]{example_matrices/Wigner_orig.pdf}
        \caption{Wigner}
        \label{subfig:wignerstem}
    \end{subfigure}
    \begin{subfigure}{0.24\linewidth}
        \includegraphics[width=1\linewidth,trim={0 0 0 0},clip]{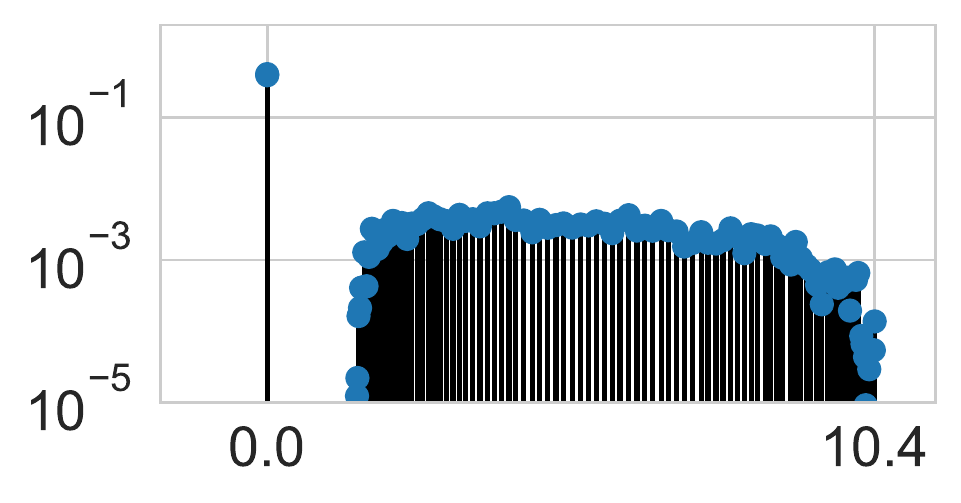}
        \caption{MP $q=6$}
        \label{subfig:mpexpamplargeq}
    \end{subfigure}
    \begin{subfigure}{0.24\linewidth}
        \includegraphics[width=1\linewidth,trim={0 0 0 0},clip]{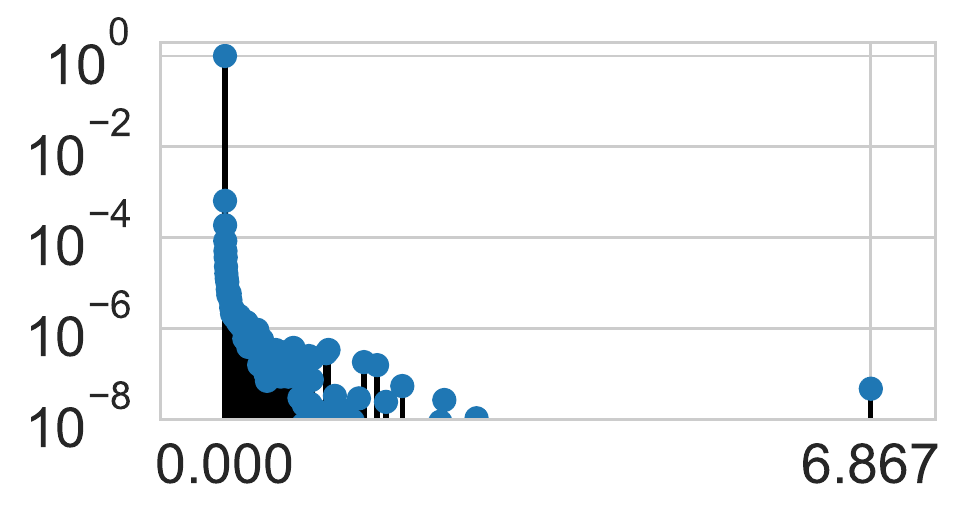}
        \caption{$\mG$ VGG-$16$ Epoch $300$}
        \label{subfig:vgg16ep300}
    \end{subfigure}
    \caption{The Generalised Gauss Newton and Residual Matrix at various Epochs of Training for the VGG-$16$ on CIFAR-$100$ and a sample Marcenko-Pastur density}\label{fig:randommatrixvsggnandres}
\end{figure}

We compare the Generalised Gauss Newton matrix of a VGG-$16$ network on the CIFAR-$100$ dataset to the Marcenko-Pastur density, which has been the subject of many comparisons \citep{pennington2017geometry,pedregosa2020average,sagun2017empirical,sagun2016eigenvalues} and the Residual matrix, assumed to be similar to the Wigner ensemble in \citet{pennington2017geometry} in Figure \ref{fig:randommatrixvsggnandres}. 
\paragraph{Where is the spectral gap?}
As can be seen from \eqref{eq:babymp}, for Marcenko-Pastur LSD when $q>1$ there is a discrete spectral component at $0$ of weight $1-1/q$. There is a spectral gap of $\sigma^{2}(1 - \sqrt{q})^{2}$ between the origin and the support of the bulk which is given by $\sigma^{2}(1\pm \sqrt{q})^{2}$. The mean of the bulk is given by $\sigma^{2}$, with an example shown in in Figure \ref{subfig:mpexpamplargeq}. 
However for the end of training for neural networks, as can be seen in Figure \ref{subfig:vgg16ep300}, not only is there no spectral gap, but the vast majority of the remaining spectral mass seems to be clustered around the origin with a rapid decay. This is in stark contrast to equation \ref{eq:babymp}, where in the case of degeneracy we expect a clear spectral gap and the maximal spectral density of the bulk to be at $\sigma^{2}$, as shown in Figure \ref{subfig:mpexpamplargeq}. The parameter $\sigma>0$ sets the scaling of the problem, how large the spectral gap is and how large we expect the bulk to extend to. $\sigma \rightarrow 0$ would solve the problem of the spectral gap, however it would predict a completely degenerate mass at the origin.

\paragraph{The Residual Matrix:}

One typical misconception in the literature is that when the optimizer reaches zero training error, the Hessian becomes the generalized Gauss Newton matrix. We showcase this explicitly in Appendix \ref{sec:zeroerror}.
As can be seen in Figures \ref{fig:randommatrixvsggnandres}, the residual similarly deviates from the proposed semi-circle density \citep{pennington2017geometry}, by displaying a sharp peak near the origin and then a sharply decaying spectral density.  Constraisting with results from \citet{papyan2018full}, we find the spectrum contains outliers, although they are not of the same scale as the GGN.
In constrast, the Wigner semi-circle, shown in Figure \ref{subfig:wignerstem}, as predicted from equation \ref{eq:babywigner} has in the log scale a reasonably uniform spectral density and a sharp cut-off. 
\paragraph{Finite sample convergence is not the problem}
For neural networks with parameter size $\mathcal{O}(10^{7})$, we would not expect significant deviation without fundamentally different underlying structure. \citet{bai1993convergence} prove the same $\mathcal{O}(P^{-1/4})$ convergence for the Marcenko-Pastur distribution, except for the special case of when $q - 1 < \epsilon$ in which case $\mathcal{O}(P^{-5/48})$. Given that in deep learning $P >> N $thus$ q \gg 1$, the same argument applies. \citet{pennington2017geometry} assume that the LSD of the residual is the Wigner semi-circle and that of the GGN is the Marcenko-Pastur and use the one to one correspondence between the LSD and Stiltjes transforms to derive the index of critical points at a given loss values. Given that we observe significant spectral deviations in the bulk, for such large matrices, it is unlikely that the conclusions reached by employing these Stiltjes transforms is valid in the setting of neural networks.


\section{Beyond elementary random matrix theory}
\label{sec:beyondrmt}
As shown in Section \ref{sec:rmtissues}, many of the claims in \citet{choromanska2015loss} are not robust to the addition of observed outliers, or substituting the GOE for a more general distribution. We further show in Figure \ref{fig:goeisabadfit} the spectra of deep neural networks deviates significantly from the Wigner semi-circle law. Other assumptions, such as the residual resembling the Wigner semi-circle and the GGN the Marcenko-Pasur, also do not hold in real neural networks. These discoveries lead us to a new questions, \textit{Do there exist random matrix ensembles which more accurately resemble neural network spectra?} 

\subsection{Products of Random Matrices}
\label{subsec:prodmatrices}
\begin{figure}[htbp]
    \centering
     \begin{subfigure}{0.24\linewidth}
        \includegraphics[width=1\linewidth,trim={0 0 0 0},clip]{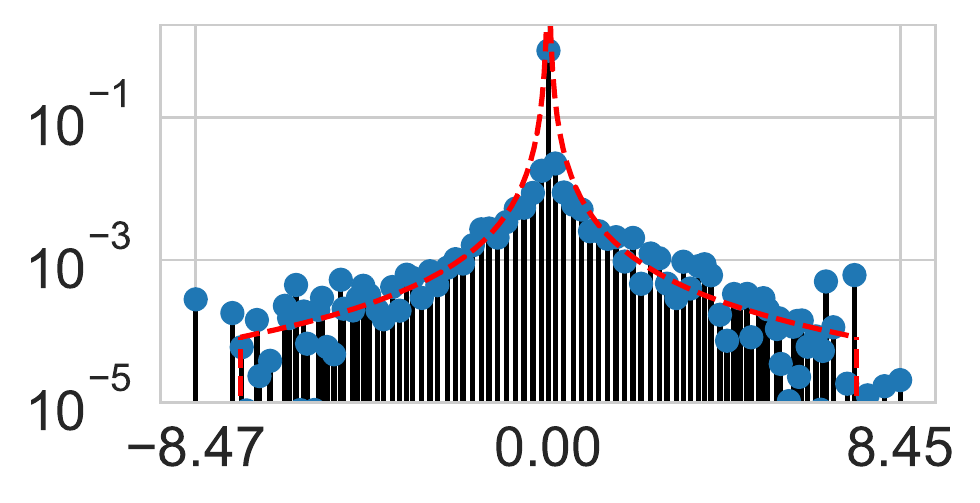}
        \caption{PWigner $L=100$}
        \label{subfig:prodwigner}
    \end{subfigure}
    \begin{subfigure}{0.24\linewidth}
        \includegraphics[width=1\linewidth,trim={0 0 0 0},clip]{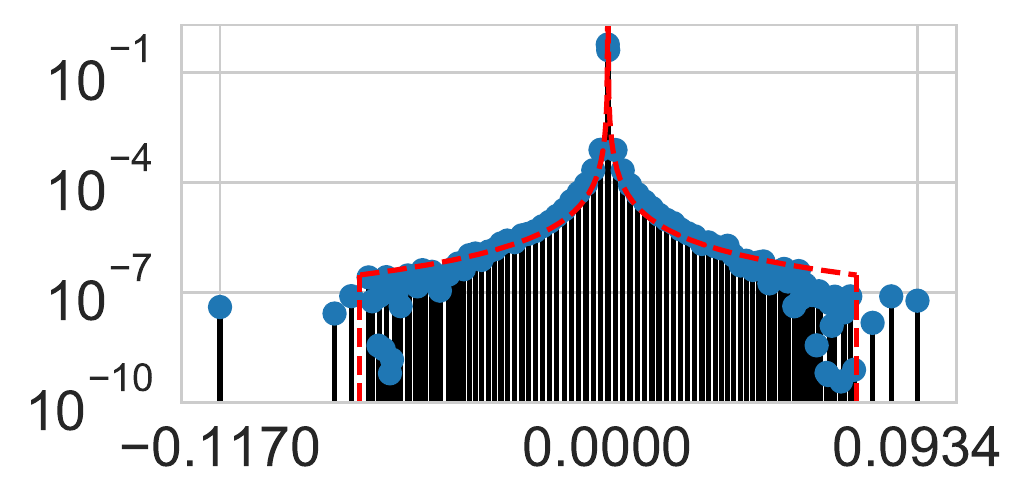}
        \caption{$\mR$ VGG-$16$ C$100$}
        \label{subfig:vgg16res}
    \end{subfigure}
    \begin{subfigure}{0.24\linewidth}
        \includegraphics[width=1\linewidth,trim={0 0 0 0},clip]{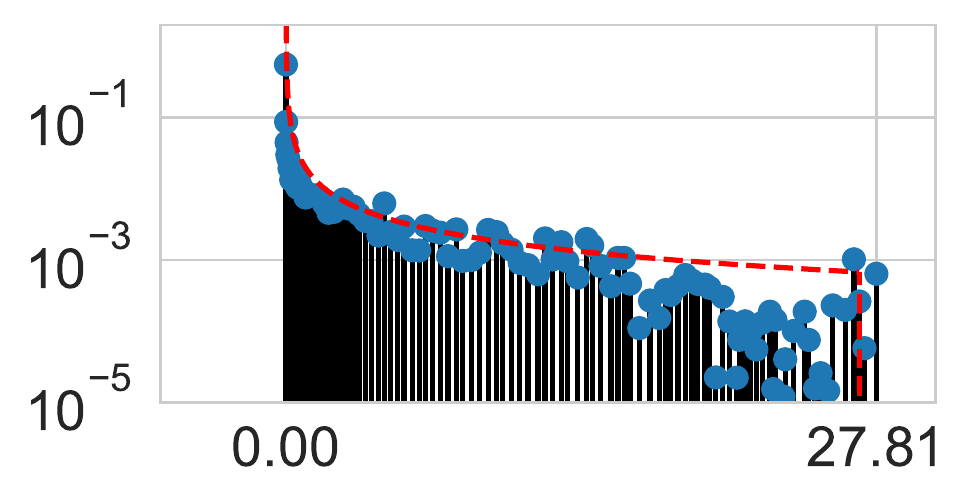}
        \caption{$\mX \in \mathbb{R}^{5000 \times 5000}$}
        \label{subfig:prodmpfullrank}
    \end{subfigure}
    \begin{subfigure}{0.24\linewidth}
        \includegraphics[width=1\linewidth,trim={0 0 0 0},clip]{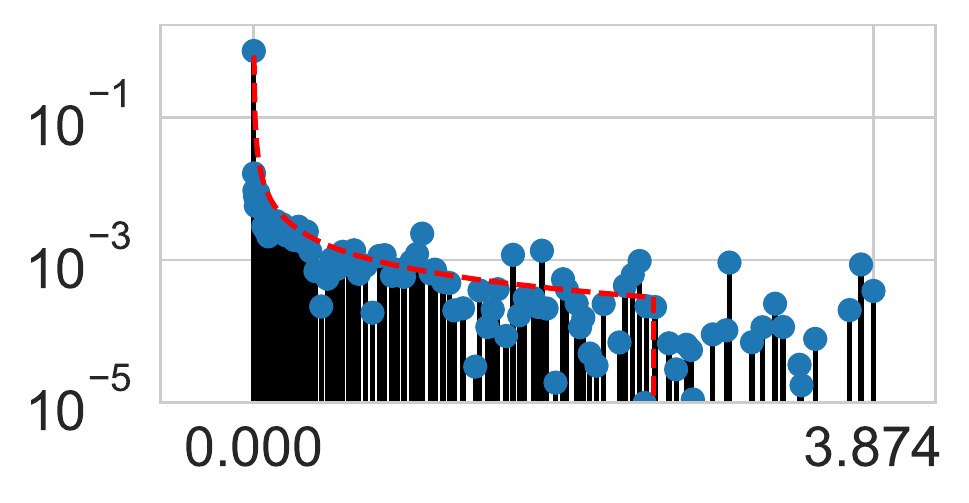}
        \caption{$\mX \in \mathbb{R}^{5000 \times 1000}$}
        \label{subfig:prodmplowrank}
    \end{subfigure}
    \caption{Spectra of Product Wigner $\mW$ and Marcenko-Pastur product ensembles, along with their theoretically predicted densities near the origin. $L$ denotes the number of products. For the Marcenko Pastur matrices $\mG = 1/T \mX\mX^{T}$, $L=5$ various dimensions of $\mX \in \mathbb{R}^{P\times T}$ were used }\label{fig:productspectra}
\end{figure}

The spectrum of a product of independent random Gaussian matrices $\mtx = \mX_{1} \times \mX_{2} ... \times \mX_{l} $ has been studied \citet{burda2010spectrum} and the spectrum shown to be given by $p(\lambda) = (1/l\pi)\sigma^{-2}|\lambda|^{-2+2/l}$ if $|\lambda| \leq \sigma$ and $0$ otherwise, where for simplicity of exposition we have taken the variance of all matrices to be equal\footnote{This can be generalised to $\sigma = \sqrt[m]{\prod_{i} \sigma_{i}}$.}. We fit this formula to the Residual matrix of the VGG-$16$ on the CIFAR-$100$ dataset in Figure \ref{subfig:vgg16res}, which we find to be in excellent agreement, we showcase the similarity between the spectrum of the Residual matrix and product of random Gaussian matrices in Figure \ref{subfig:prodwigner}. We plot simulated examples of the corresponding Marcenko-Pastur equivalent $\mtx \mtx^{T}$, where we allow the dimensions of $\mtx$ to vary in Figure \ref{subfig:prodwigner}. We find that the spectra produced, shown in Figures \ref{subfig:prodmplowrank} and \ref{subfig:prodmpfullrank} are in excellent agreement with the typical spectral shapes observed in the GGN of deep neural network Hessians (as shown in section \ref{subsec:ggn}) and with the theoretically predicted spectral density near the origin, which goes as $\lambda^{-(1+\frac{1}{s})}$ in the vicinity of $\lambda \approx 0$. 

\paragraph{Intra-connected Minima and Matrix dimensions:}

One interesting observation about deep learning loss surfaces is that many areas of low loss can be connected with simple curves \citep{garipov2018loss,draxler2018essentially}, this has been used to propose fast geometric ensembling and has been used for justification for averaging weights during training, giving improved test performance \citep{izmailov2018averaging} and recently integrated into adaptive optimisers to achieve state of the art test performance in typical image problems \citep{granziol2020iterate}. For such curves to exist, an appreciable amount of spectral mass at every point in weight space must contain a large proportion of zero or very near zero eigenvalues. This is appreciably not the case for the Wigner semi-circle or any free convolution between the wigner semi-circle and the Marcenko-Pastur. However for both the Wigner product and the Wishart product equivalent, the predicted spectrum diverges around the origin, giving precisely this effect. Moreover, when the dimensions of the matrices multiplied vary greatly in size, we actually recover a discrete component at the origin. This can be seen as the non-holomorphic M-transform $M_{P}(z,\Tilde{z})$ solves the $L$'th order polynomial equation, where $R_{l} = N_{l}/N_{L}$ is the ratio of the dimension of $l$'th and last matrices $\mX \in \mathbb{R}^{N_{l} \times N_{l+1}}$, $\prod_{i}^{L}\bigg(\frac{M_{\mP}(z,\hat{z})}{R_{i}}+1\bigg) = \frac{|z|^{2}}{\sigma^{2}}$
, which can be easily solved for $L=2$ and from the definition of the Green's function $G_{\mP}(z,\hat{z}) = (M+1)/z$, using the Stiltjes-Perron identity we derive the spectral density in the complex space for $|z| \leq \sigma$
\begin{equation}
    \rho(z,\hat{z}) = \frac{R}{\pi\sigma^{2}\sqrt{(1-R)^{2}+4R|z|^{2}/\sigma^{2}}} + (1-R)\delta(z,\hat{z})
\end{equation}
Although there is no systematic way to solve an $L$'th order polynomial for $L \geq 2$, under the assumption that $R_{i} \gg 1$ we can expand the to second order terms in $1/R_{i}$, further assuming that $\langle 1/R^{2}\rangle = \langle 1/R \rangle^{2}$, we arrive at a degeneracy fraction
$1- \frac{\langle 1/R \rangle}{L}$. Hence for many multiplications (large $L$) and a large ratio between the multiplicative matrices, the degeneracy increases.

\subsection{Percolated Hessian Spectra}
\label{subsec:sparsematrices}

\begin{figure}[htbp]
    \centering
     \begin{subfigure}{0.24\linewidth}
        \includegraphics[width=1\linewidth,trim={0 0 0 0},clip]{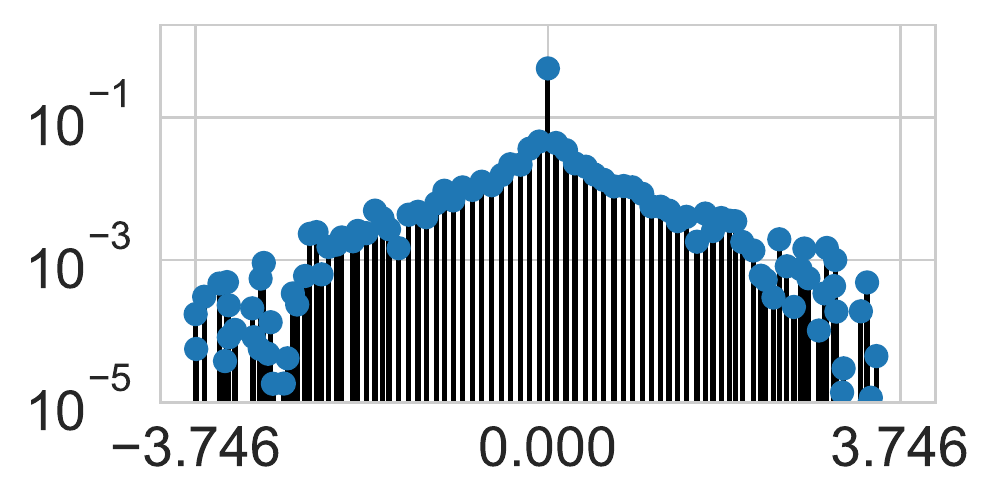}
        \caption{S-Wigner $k=1$}
        \label{subfig:sparsewigner}
    \end{subfigure}
    \begin{subfigure}{0.24\linewidth}
        \includegraphics[width=1\linewidth,trim={0 0 0 0},clip]{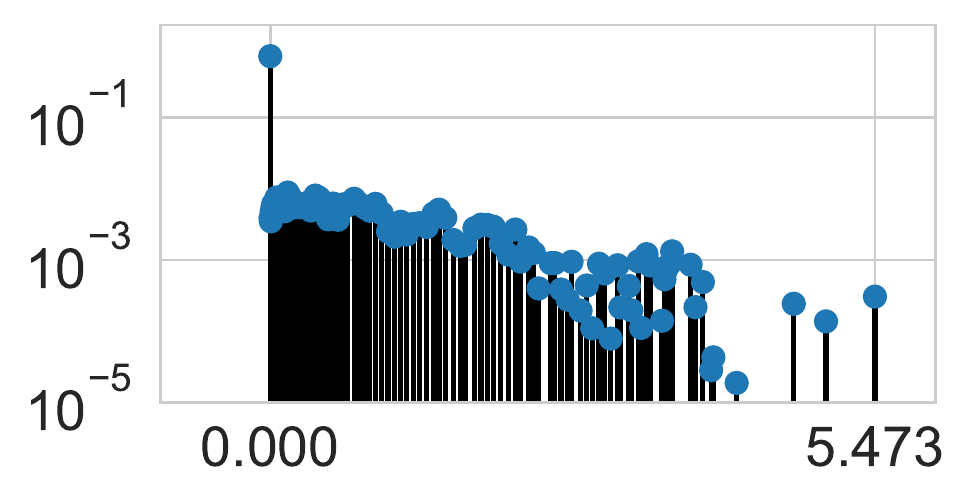}
        \caption{S-MP $k=1$}
        \label{subfig:sparsempgraph}
    \end{subfigure}
    \begin{subfigure}{0.24\linewidth}
        \includegraphics[width=1\linewidth,trim={0 0 0 0},clip]{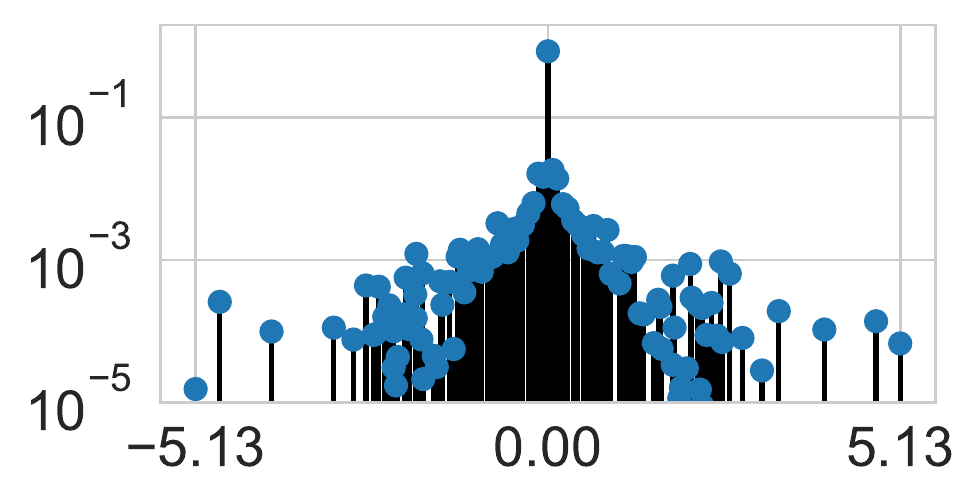}
        \caption{SP-Wigner $k \& L=2$}
        \label{subfig:sparsewignerpkl22}
    \end{subfigure}
    \begin{subfigure}{0.24\linewidth}
        \includegraphics[width=1\linewidth,trim={0 0 0 0},clip]{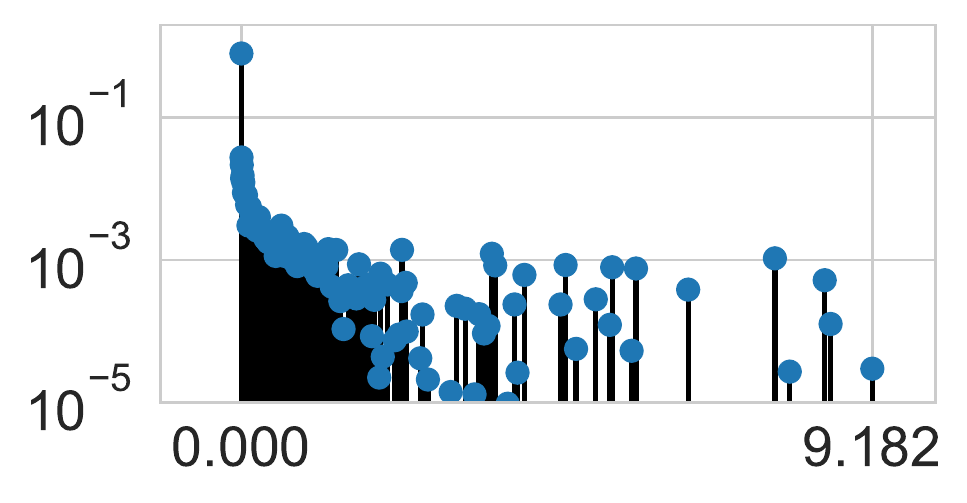}
        \caption{SP-MP $k=1, L=2$}
        \label{subfig:sparsempgraphkl12}
    \end{subfigure}
    \caption{Spectra of Sparse Wigner $\mW$ and Sparse Wigner product $\mW_{1} \prod \mW_{2}$ ensembles along with Sparse Marcenko-Pastur and Sparse Marcenko-Pastur product ensembles. $k$ denotes the sparsity constant, and $L$ denotes the number of products}\label{fig:sparsespectra}
\end{figure}
\begin{figure}[htbp]
    \centering
     \begin{subfigure}{0.24\linewidth}
        \includegraphics[width=1\linewidth,trim={0 0 0 0},clip]{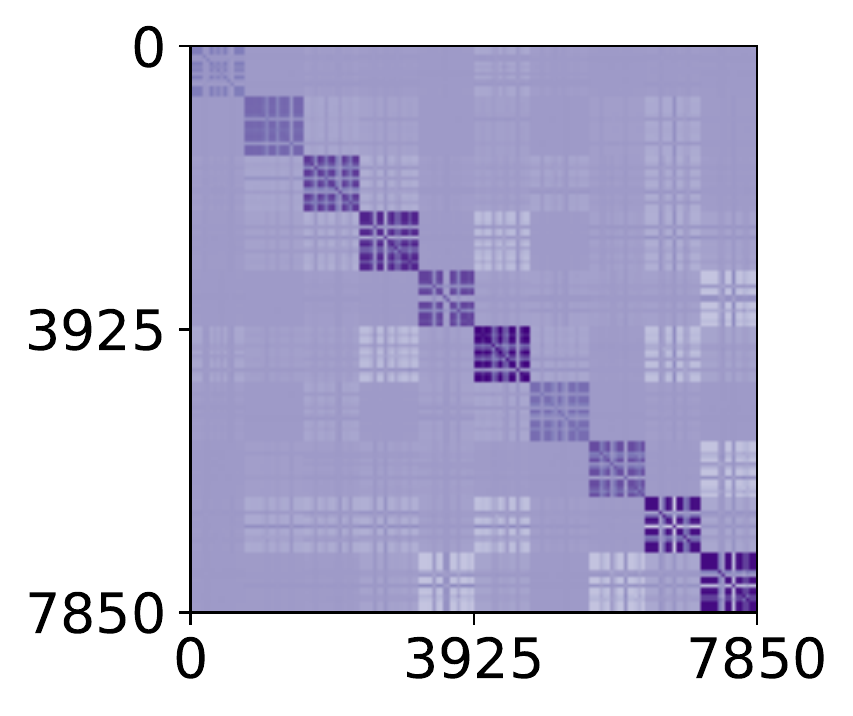}
        \caption{$\mH_{logistic}$ Heat Map}
        \label{subfig:logisticheatmap}
    \end{subfigure}
     \begin{subfigure}{0.22\linewidth}
        \includegraphics[width=1\linewidth,trim={0 0 0 0},clip]{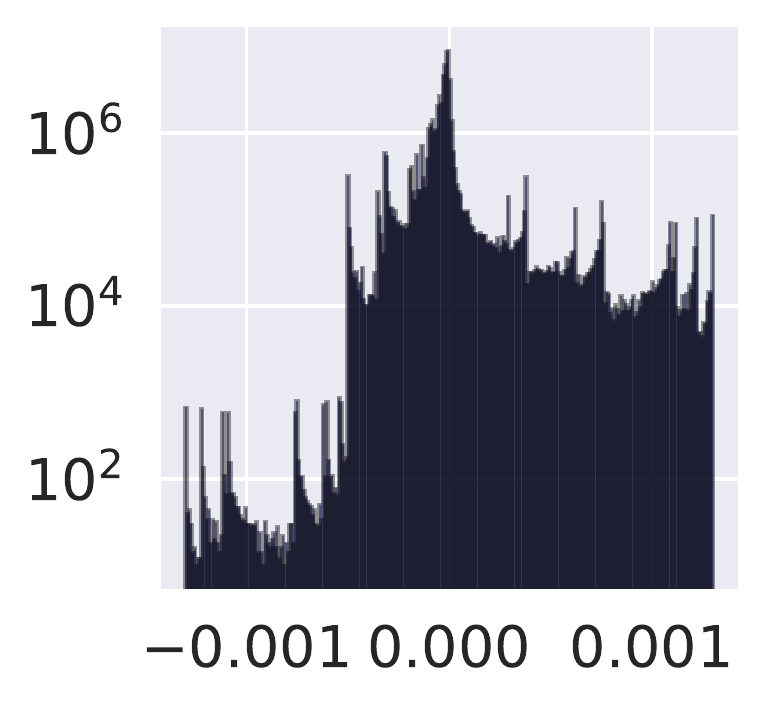}
        \caption{$\mH_{logistic}[i,j]$}
        \label{subfig:logisticelements}
    \end{subfigure}
    \begin{subfigure}{0.24\linewidth}
        \includegraphics[width=1\linewidth,trim={0 0 0 0},clip]{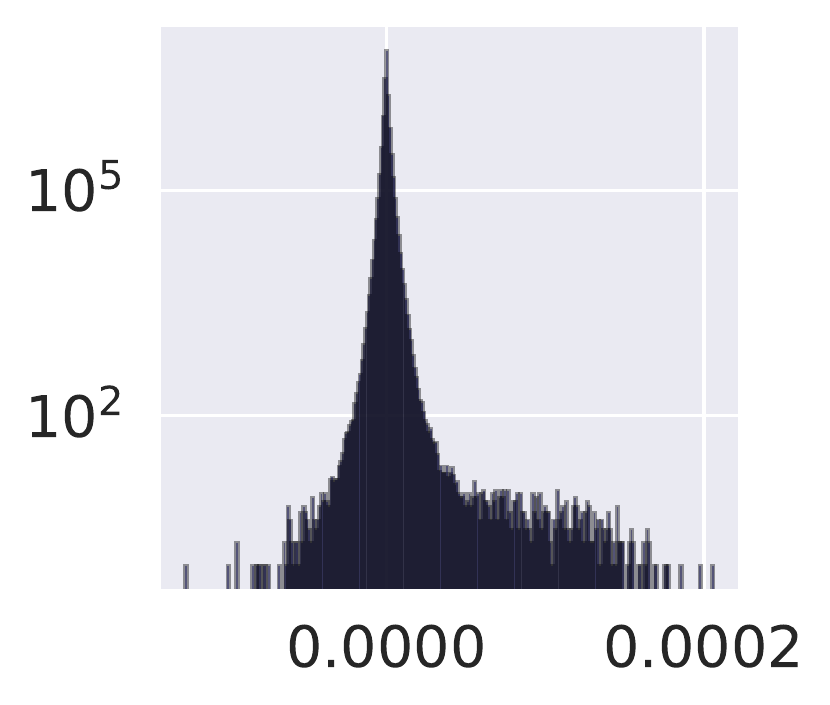}
        \caption{$\mH_{VGG16}[i=1,j]$}
        \label{subfig:vgg16elements}
    \end{subfigure}
        \begin{subfigure}{0.24\linewidth}
        \includegraphics[width=1\linewidth,trim={0 0 0 0},clip]{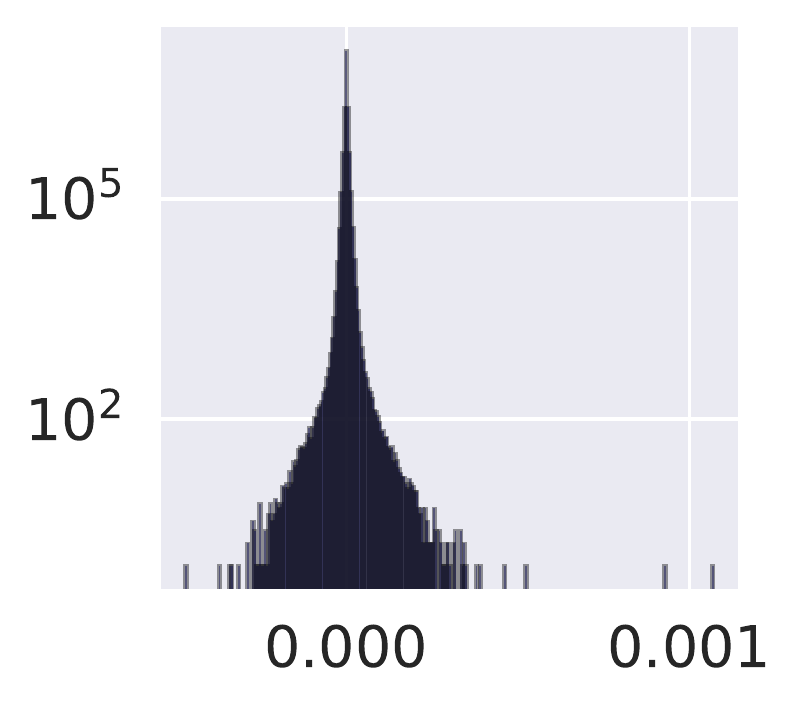}
        \caption{$\mH_{VGG16BN}[i=1,j]$}
        \label{subfig:vgg16bnelements}
    \end{subfigure}
    \caption{Heatmap and histogram of the Hessian of Logistic regression on MNIST and rows from the VGG-$16$/BN on CIFAR-$100$}\label{fig:issparsityrelevant}
\end{figure}

The idea, of randomly removing connections in graphs, is known as \textit{percolation} in the Physics community and is used to study impure conductors \citep{evangelou1983quantum}. We consider the simplest sparsification process, where we set elements a Wigner random matrix model to $0$ with probability $1-p$, independently of one another. For the sparsity to have an appreciable effect on the spectrum $p = k/P$, this can be proved rigorously \citep{tao2012topics,khorunzhy2001sparse,chung2009giant}. We show the spectra of sparsified Wigner and Marcenko-Pastur matrices in Figure \ref{fig:sparsespectra}. We note that both the sparse Wigner and Marcenko-Pastur matrices, Figures \ref{subfig:sparsewigner} and \ref{subfig:sparsempgraph} fit the empircal observations much better than their non sparse counterparts. Interestingly the sparse product spectra, shown in Figures \ref{subfig:sparsewignerpkl22} and \ref{subfig:sparsempgraphkl12} seem even more convincing.
Moments for percolated ER graphs have been derived \citep{bauer2001random}, the spectrum always contains a separated discrete component and the existence of a sharp spectral peak at the origin. \citet{rodgers1988density} show that the distribution has tails extending beyond the semi circle $p(\lambda) \approx (ep/\lambda^{2})^{\lambda^{2}}$, where $p$ is the number of non zero elements per row. \citet{benaych2019largest} show a crossover in the behaviour of extreme eigenvalues when the constant of proportionality $k \approx \log P$. 
For sparse Erdos Renyi graphs $\mathcal{G}(P,k/P)$, for $k \ll \log P$, $P^{1-\mathcal{O}(1)}$ extremal eigenvalues escape the support of the semi-circle and follow the distribution \citep{benaych2019largest}
$
    \lambda_{k} \approx \sqrt{\frac{\log (P/k)}{\log((\log P)/d)}}, \thinspace k \geq P^{1-\epsilon}, \thinspace \epsilon \in (0,1)
$
which also grows with matrix dimension and hence cannot be controlled by the use of the large deviation principle as argued in Section \ref{subsec:GOEismagical}. To evaluate the extent of sparsity in deep learning, we run Logistic Regression on the MNIST dataset, which with parameter count $P=7850$, visualising the Hessian in heatmap form in Figure \ref{subfig:logisticheatmap} and as a histogram in Figure \ref{subfig:logisticelements}. We see from both figures that although there is a large proportion of near zero elements, there is appreciable mass across the support. For the VGG-$16$ architecture on CIFAR-$100$ with and without batch normalization with $P = [15299748,15291300]$ we sample rows from the relevant matrix, shown in Figures \ref{subfig:vgg16elements} and \ref{subfig:vgg16bnelements}. Although the shape of the distribution for each row varies, there is considerable mass outside the origin. For the VGG-$16$ only $8 \pm 2 \%$ of the sampled elements are less than $10^{-8}$ (the GPU precision threshold) and $84 \pm 13 \%$ as less than $10^{-6}$. This evidences that sparsity is a relevant factor in neural network spectra, but not sufficient to determine spectral characteristics alone. 


\section{Conclusion}
This paper evaluates the Hessian spectral densities of modern neural networks, such as the VGG-$16$ and PreResNet-$110$ on the CIFAR-$100$ dataset. It is found that they and their decompositions (such as the GGN and Residual) exhibit significant deviations from theoretically employed Wigner and Wishart esnembles. We further show that the layered loss surface as a function of energy \citep{choromanska2015loss}, is not robust to removing the assumption of identical Gaussianity in the proofs, or the addition of outliers (which are observed in practice). Whilst we do not go over all the examples in the literature, we conjecture that similar results will hold for proof techniques using the Stiltjes transforms corresponding to the Marcenko-Pastur and Wigner ensembles, in the context of deep learning. Alternative random matrix ensembles, such as Wiger/Wishart products or percolated ensembles are investigated. Both ensembles are shown to significantly better fit the observed spectra and could form promising lines of further research. They both also predict rank dengeracy at and near the origin, which fits with experimental evidence that low loss regions can be easily connected by simple curves. We investigate the nature of the dependence of the degeneracy as a function of layer width and depth and show under certain assumptions that different layer widths and an increase in depth increases the degeneracy factor.

\section{Broader Impact}
This paper is primarily theoretical and we expect the impact to be mainly limited to aiding the understanding of Deep Neural Networks to a community of researchers and practitioners. One theoretical impact is that spin glass models and the associated Gaussian orthogonal ensemble model of the Hessian are not useful for modelling real neural networks. As with any evidence that an established theoretical framework is not appropriate, there is a risk of decreased confidence in the use of and perceived confidence in understanding of neural networks. However, even though we see this work as having a negative impact on current random matrix theory in deep learning work, we see our inclusion of product and percolated ensembles and the extensive experimental agreement as a major step forward in a more complete understanding of neural network loss surfaces. This could have major implications for understanding generalisation and optimisation for these networks. Given the held out test set performance of neural networks across a plethora of tasks, we see any fundamental research, especially research which challenges underlying assumptions and commonly held beliefs in a novel and rigorous manner as impactful and helpful. We believe our research falls under this category and is of help to the community. 
\bibliography{example_paper}
\bibliographystyle{plainnat}

\newpage
\appendix

\section{Experiment Details}
\label{sec:experimentdetails}
\subsection{Image Classification Experiments} 

\paragraph{Hyperparameter Tuning} For SGD for the VGG-$16$ without batch noramlisation we use a learning rate of $\alpha=0.05$, for any networks with batch-normalisation we use $\alpha =0.1$, all experiments have a weight decay of $\gamma = 0.0005$. 

\subsection{Experimental Details} 
\label{sec:expdetails}
For all experiments,  we use the following learning rate schedule for the learning rate at the $t$-th epoch:
\begin{equation}
    \alpha_t = 
    \begin{cases}
      \alpha_0, & \text{if}\ \frac{t}{T} \leq 0.5 \\
      \alpha_0[1 - \frac{(1 - r)(\frac{t}{T} - 0.5)}{0.4}] & \text{if } 0.5 < \frac{t}{T} \leq 0.9 \\
      \alpha_0r, & \text{otherwise}
    \end{cases}
\end{equation}
where $\alpha_0$ is the initial learning rate. In the motivating logistic regression experiments on MNIST, we used $T = 50$. $T = 300$ is the total number of epochs budgeted for all CIFAR experiments. We set $r = 0.01$ for all experiments. 

\section{Lanczos algorithm}
\label{sec:lanczos}
In order to empirically analyse properties of modern neural network spectra with tens of millions of parameters $N = \mathcal{O}(10^{7})$, we use the Lanczos algorithm \citep{meurant2006lanczos}, provided for deep learning by \citet{granziol2019mlrg}. It requires Hessian vector products, for which we use the Pearlmutter trick \citep{pearlmutter1994fast} with computational cost $\mathcal{O}(NP)$, where $N$ is the dataset size and $P$ is the number of parameters. Hence for $m$ steps the total computational complexity including re-orthogonalisation is $\mathcal{O}(NPm)$ and memory cost of $\mathcal{O}(Pm)$. In order to obtain accurate spectral density estimates we re-orthogonalise at every step \citep{meurant2006lanczos}. We exploit the relationship between the Lanczos method and Gaussian quadrature, using random vectors to allow us to learn a discrete approximation of the spectral density.  A quadrature rule is a relation of the form,
	\begin{equation}
		\label{eq:quadraturerule}
		\int_{a}^{b}f(\lambda)d\mu(\lambda) = \sum_{j=1}^{M}\rho_{j}f(t_{j})+R[f]
	\end{equation}
for a function $f$, such that its Riemann-Stieltjes integral and all the moments exist on the measure $d\mu(\lambda)$, on the interval $[a,b]$ and where $R[f]$ denotes the unknown remainder. The nodes $t_{j}$ of the Gauss quadrature rule are given by the Ritz values and the weights (or mass) $\rho_{j}$ by the squares of the first elements of the normalized eigenvectors of the Lanczos tri-diagonal matrix \citep{golub1994matrices}. The main properties of the Lanczos algorithm are summarized in the theorems \ref{theorem:lanczoseigenvalues},\ref{theorem:lanczosspectrum}
\begin{theorem}
	\label{theorem:lanczoseigenvalues}
	Let $H^{N\times N}$ be a symmetric matrix with eigenvalues $\lambda_{1}\geq .. \geq \lambda_{n}$ and corresponding orthonormal eigenvectors $z_{1},..z_{n}$. If $\theta_{1}\geq .. \geq \theta_{m}$ are the eigenvalues of the matrix $T_{m}$ obtained after $m$ Lanczos steps and $q_{1},...q_{k}$ the corresponding Ritz eigenvectors then
	\begin{equation}
	\begin{aligned}
	& \lambda_{1} \geq \theta_{1} \geq \lambda_{1} - \frac{(\lambda_{1}-\lambda_{n})\tan^{2}(\theta_{1})}{(c_{k-1}(1+2\rho_{1}))^{2}} \\
	& \lambda_{n} \leq \theta_{k} \leq \lambda_{m} + \frac{(\lambda_{1}-\lambda_{n})\tan^{2}(\theta_{1})}{(c_{k-1}(1+2\rho_{1}))^{2}} \\
	\end{aligned}	
	\end{equation}
	where $c_{k}$ is the chebyshev polyomial of order $k$
\end{theorem}
Proof: see \citep{golub2012matrix}.
\begin{theorem}
	\label{theorem:lanczosspectrum}
	The eigenvalues of $T_{k}$ are the nodes $t_{j}$ of the Gauss quadrature rule, the weights $w_{j}$ are the squares of the first elements of the normalized eigenvectors of $T_{k}$
\end{theorem}
Proof: See \citep{golub1994matrices}. The first term on the RHS of \eqref{eq:quadraturerule} using Theorem \ref{theorem:lanczosspectrum} can be seen as a discrete approximation to the spectral density matching the first $m$ moments $v^{T}H^{m}v$ \citep{golub1994matrices,golub2012matrix}, where $v$ is the initial seed vector. Using the expectation of quadratic forms, for zero mean, unit variance random vectors, using the linearity of trace and expectation
\begin{equation}
\begin{aligned}
\mathbb{E}_{v}\text{Tr}(v^{T}H^{m}v) & =  \text{Tr}\mathbb{E}_{v}(vv^{T}H^{m}) = \text{Tr}(H^{m})
 = \sum_{i=1}^{N}\lambda_{i} = N \int_{\lambda \in \mathcal{D}} \lambda d\mu(\lambda) \\
\end{aligned}
\end{equation}
The error between the expectation over the set of all zero mean, unit variance vectors $v$ and the monte carlo sum used in practice can be bounded \citep{hutchinson1990stochastic,roosta2015improved}. However in the high dimensional regime $N \rightarrow \infty$, we expect the squared overlap of each random vector with an eigenvector of $H$, $|v^{T}\phi_{i}|^{2} \approx \frac{1}{N} \forall i$, with high probability. This result can be seen by computing the moments of the overlap between Rademacher vectors, containing elements $P(v_{j} = \pm 1) = 0.5$. Further analytical results for Gaussian vectors have been obtained \citep{cai2013distributions}.

\subsection{Wigner Ensemble \& the Semi-Circle Law}
\label{subsec:wignerandgoe}
One of the most important eigenvalue densities in the study of random matrices, akin to the normal distribution for multivariate statistics, is the Wigner semi-circle. Wigner matrices are defined in Definition \ref{def:wignerensemble} and the Semi-circle distribution result in Theorem \ref{theorem:semicircle}. 
\begin{definition}{}
\label{def:wignerensemble}
Let $\{Y_{i}\}$ and $\{Z_{ij}\}_{1\leq i\leq j}$ be two real-valued families of zero mean, i.i.d random variables, Furthermore suppose that $\mathbb{E}Z_{12}^{2}=1$ and for each $k \in \mathbb{N}$
\begin{equation}
    \max(E|Z_{ij}^{k}|,E|Y_{i}|^{k})< \infty
\end{equation}
Consider a $P \times P$ symmetric matrix $\mM_{P}$, whose entries are given by
\begin{equation}
  \begin{cases}
    \mM_{P}(i,i) = Y_{i}\\
    \mM_{P}(i,j) = Z_{ij} = \mM_{P}(j,i) ,& \text{if } x\geq 1 
\end{cases} 
\end{equation}
The Matrix $\mM_{P}$ is known as a real symmetric Wigner matrix.
\end{definition}
\begin{theorem}
\label{theorem:semicircle}
Let $\{M_{P}\}^{\infty}_{n=1}$ be a sequence of Wigner matrices, and for each $n$ denote $X_{P} = \mM_{P}/\sqrt{P}$. Then $\mu_{X_{P}}$, converges weakly, almost surely to the semi-circle  distribution,
\begin{equation}
    \sigma(x)dx = \frac{1}{2\pi}\sqrt{4-x^{2}}\mathbf{1}_{|x|\leq 2}
\end{equation}
\end{theorem}
\paragraph{The Gaussian Orthogonal Ensemble (GOE)} is a special case of the Wigner matrix for which the entries are Gaussian. The GOE is invariant under rotation, which allows us to determine analytical forms for the eigenvalue perturbations.
\subsection{The Marcenko-Pastur Law}
\label{subsec:mplaw}
An equally important limiting law for the limiting spectral density of many classes of matrices constrained to be positive definite, such as covariance matrices is the Marcenko-Pastur law \citep{marvcenko1967distribution}. Formally, given a matrix $\mX \in \mathbb{R}^{P\times T}$ with i.i.d zero mean entires with variance $\sigma^{2}<\infty$. Let $\lambda_{1}\geq \lambda_{2},...\geq\lambda_{P}$ be eigenvalues of $\mY_{n} = \frac{1}{T}\mX\mX^{T}$. The random measure $\mu_{P}(A) = \frac{1}{P}\#\{\lambda_{j} \in A\}$, $A \in \mathbb{R}$
\begin{theorem}
\label{theorem:mplaw}
Assume that $P,N \rightarrow \infty$ and the ratio $P/N \rightarrow q \in (0,\infty)$ (this is known as the Kolmogorov limit) then $\mu_{P}\rightarrow \mu$ in distribution where
\begin{equation}
  \begin{cases}
    (1-\frac{1}{q})\mathbbm{1}_{0\in A}+\nu_{1/q}(A) ,& \text{if } q> 1 \\
    \nu_{q}(A) ,& \text{if } 0\leq q \leq 1 
\end{cases} 
\end{equation}
\begin{equation}
\begin{aligned}
    & d\nu_{q} = \frac{1}{2\pi\sigma^{2}}\frac{\sqrt{(\lambda_{+}-x)(x-\lambda_{-}}}{\lambda x}\\
    & \lambda_{\pm} = \sigma^{2}(1\pm \sqrt{q})^{2}
    \end{aligned}
\end{equation}
\end{theorem}
\section{Zero training error is not zero loss}
\label{sec:zeroerror}

The cross entropy loss \eqref{eq:crossentropy} is only zero when the model probability of the correct class is $1$. The gradient of the softmax is only zero when $z_{i} \rightarrow \infty$, i.e it is not sufficient to have $0$ training error, but we must also have $0$ training loss, which is only possible with infinite network outputs into the softmax. 

We showcase this  on the VGG-$16$ network \citep{simonyan2014very} on the CIFAR-$10$ dataset with no data-augmentation. 
As shown in Figures \ref{subfig:trainacccif10notrans}, \ref{subfig:trainlosscif10notrans} we achieve $100 \%$ training accuracy and $\approx 10^{-3}$ loss. We plot the spectral densities of the Hessian and Generalised Gauss Newton, in Figures \ref{subfig:cif10notranshess} and \ref{subfig:cif10notransggn} respectively using \citet{granziol2019mlrg}. The underlying Lanczos algorithm gives a lower/upper bound for the largest/smallest extremal eigenvalues \citep{meurant2006lanczos,granziol2019mlrg} and hence the smallest eigenvalue of the Hessian is upper bounded by $-0.004$. Hence the Hessian is not positive definite, even at $0$ error and does not equal the generalised Gauss-Newton.

For the cross-entropy loss, the most common loss function in Deep Learning, this means the residual matrix cannot be neglected such as in \citet{sagun2016eigenvalues} and hence arguments pertaining to the rank or positive definite nature of the Hessian, depend heavily on the nature of the Residual matrix. We hence specifically investigate the Residual matrix and its evolution during training.

\begin{table}[]
\vspace{-12pt}
\begin{tabular}{lc}
\begin{minipage}{.48\textwidth}
\begin{figure}[H]
    \centering
     \begin{subfigure}{0.47\linewidth}
        \includegraphics[width=1\linewidth]{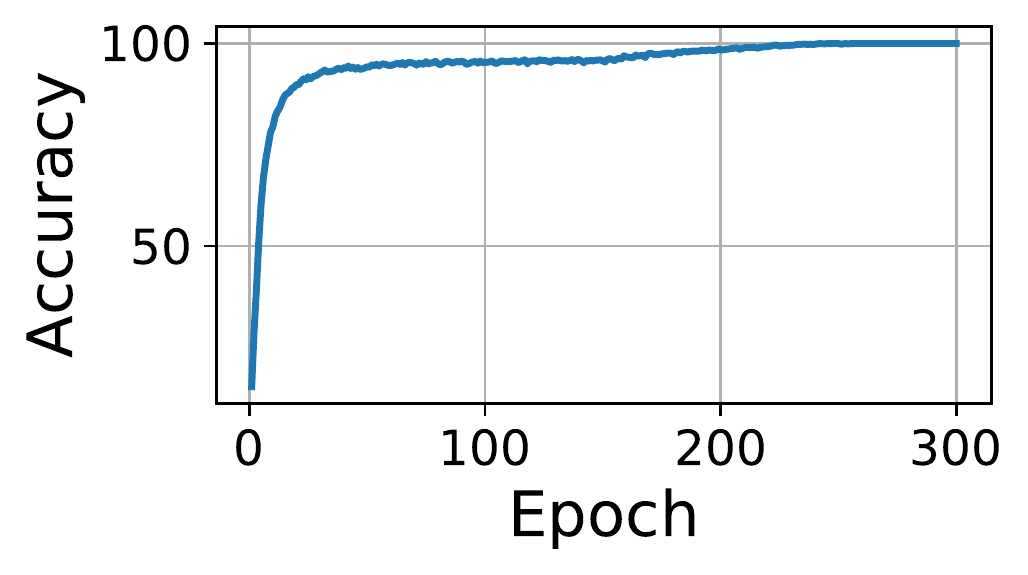}
        \caption{Train Accuracy}
        \label{subfig:trainacccif10notrans}
    \end{subfigure}
    \hspace{0.02\linewidth}
    \begin{subfigure}{0.47\linewidth}
        \includegraphics[width=1\linewidth]{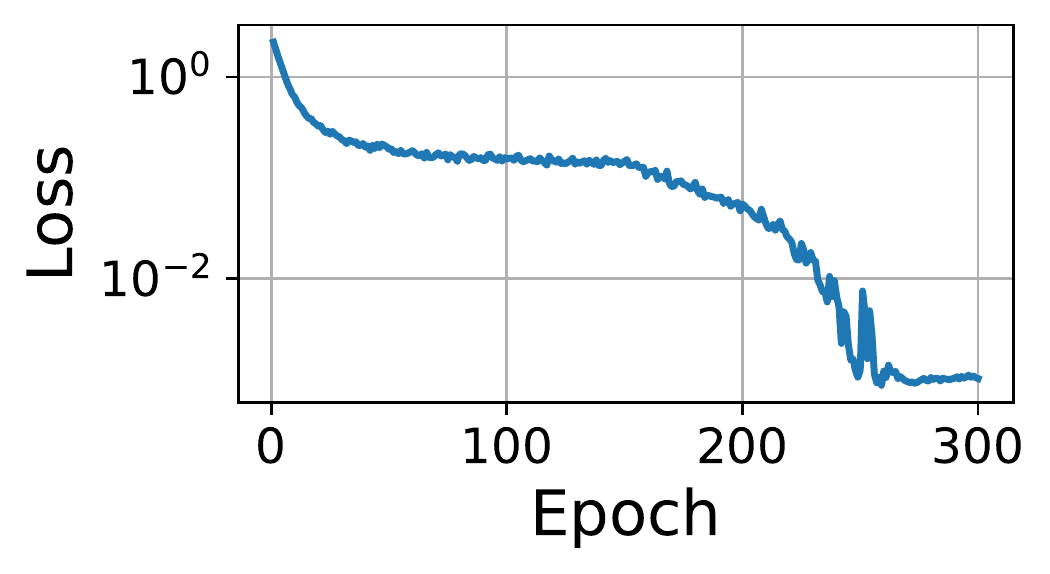}
        \caption{Train Loss}
        \label{subfig:trainlosscif10notrans}
    \end{subfigure}
    \caption{CIFAR-$10$ VGG-$16$ training curves}\label{fig:cif10notranscurves}
\end{figure}
\end{minipage}
\begin{minipage}{.48\textwidth}
\begin{figure}[H]
    \centering
     \begin{subfigure}{0.47\linewidth}
        \includegraphics[width=1\linewidth]{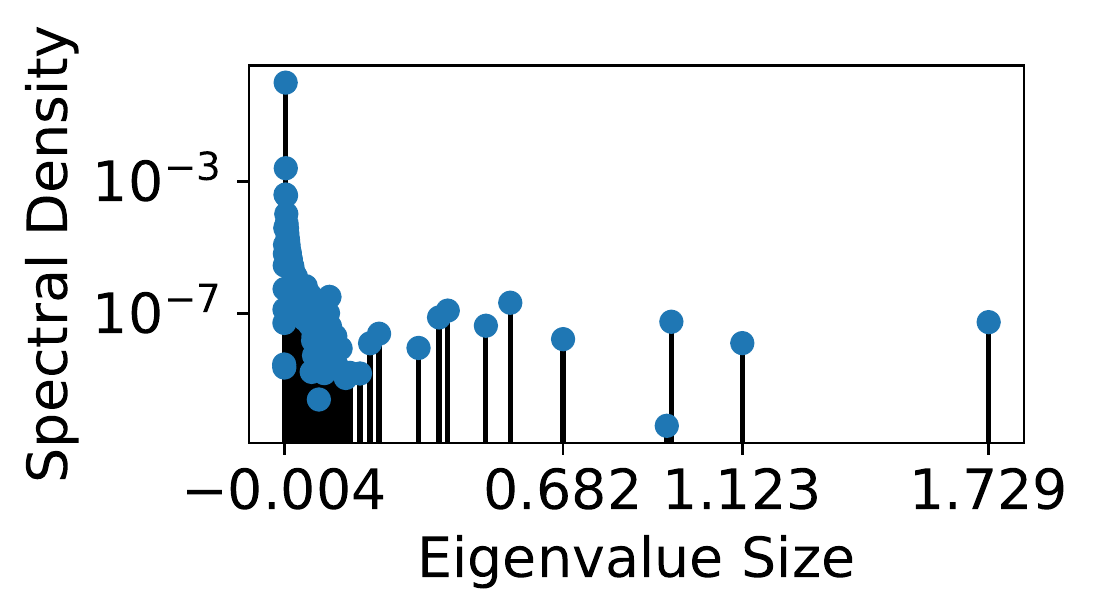}
        \caption{Hessian}
        \label{subfig:cif10notranshess}
    \end{subfigure}
    \hspace{0.02\linewidth}
    \begin{subfigure}{0.47\linewidth}
        \includegraphics[width=1\linewidth]{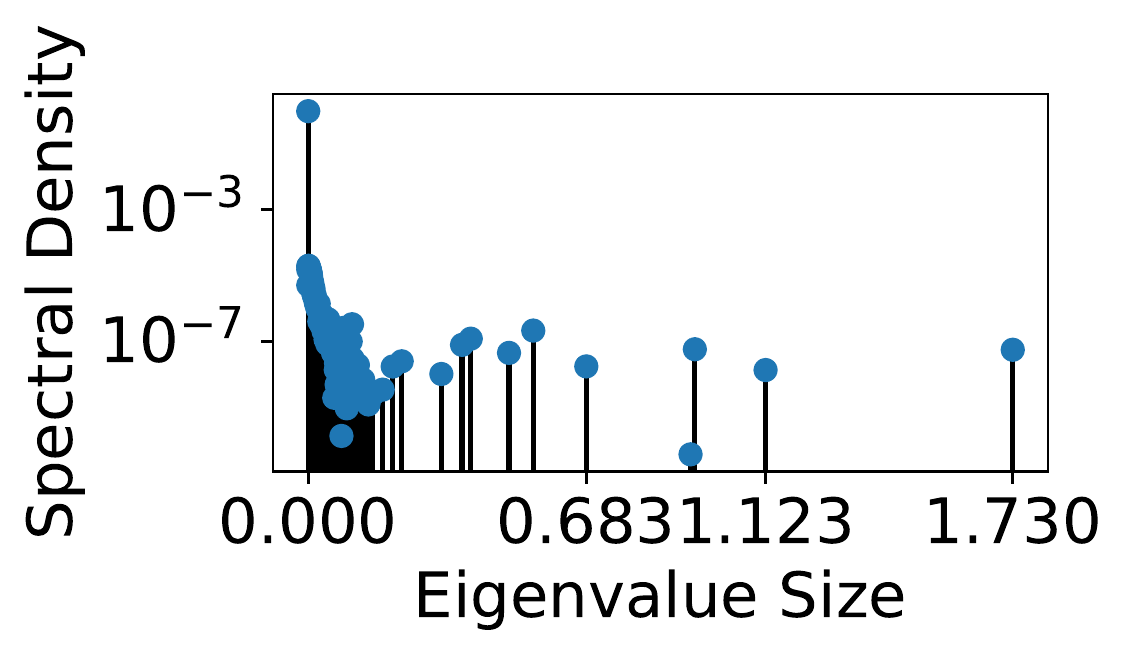}
        \caption{GGN}
        \label{subfig:cif10notransggn}
    \end{subfigure}
    \caption{CIFAR-$10$ VGG-$16$ Hessian Spectra}\label{fig:cif10notransspec}
\end{figure}
\end{minipage}
\end{tabular}
\vspace{-12pt}
\end{table}	
\section{The spiked GOE}
\label{sec:beastderivation}

Following the notation of \citep{bun2017cleaning} the resolvent of a matrix $H$ is defined as
\begin{equation}
    G_{H}(z) = (zI_{N}-H)^{-1}
\end{equation}
with  $z = x + i\eta \in \mathbb{C}$. The normalised trace operator of the resolvent, in the $N \rightarrow \infty$ limit
\begin{equation}
   \mathcal{S}_{N}(z) = \frac{1}{N}\text{Tr}[G_{H}(z)] \xrightarrow{N \rightarrow \infty} \mathcal{S}(z) = \int \frac{\rho(u)}{z-u}du
\end{equation}
is known as the Stieltjes transform of $\rho$. The functional inverse of the Siteltjes transform, is denoted the blue function $\mathcal{B}(\mathcal{S}(z)) = z$. The R transform is defined as 
\begin{equation}
    \mathcal{R}(w) = \mathcal{B}(w) - \frac{1}{w}
\end{equation}
crucially for our calculations, it is known that the $\mathcal{R}$ transform of the Wigner ensemble is
\begin{equation}
    \mathcal{R}_{W}(z) = \sigma^{2}z
\end{equation}
\begin{definition}{}
Let $\{Y_{i}\}$ and $\{Z_{ij}\}_{1\leq i\leq j}$ be two real-valued families of zero mean, i.i.d random variables, Furthermore suppose that $\mathbb{E}Z_{12}^{2}=1$ and for each $k \in \mathbb{N}$
\begin{equation}
    max(E|Z_{12}^{k},E|Y_{1}|^{k})< \infty
\end{equation}
Consider an $n \times n$ symettric matrix $M_{n}$, whose entries are given by
\begin{equation}
  \begin{cases}
    M_{n}(i,i) = Y_{i}\\
    M_{n}(i,j) = Z_{ij} = M_{n}(j,i) ,& \text{if } x\geq 1 
\end{cases} 
\end{equation}
The Matrix $M_{n}$ is known as a real symmetric Wigner matrix.
\end{definition}
\begin{theorem}
Let $\{M_{n}\}^{\infty}_{n=1}$ be a sequence of Wigner matrices, and for each $n$ denote $X_{n} = M_{n}/\sqrt{n}$. Then $\mu_{X_{n}}$, converges weakly, almost surely to the semi circle  distribution,
\begin{equation}
    \sigma(x)dx = \frac{1}{2\pi}\sqrt{4-x^{2}}\mathbf{1}_{|x|\leq 2}
\end{equation}
\end{theorem}
the property of freeness for non commutative random matrices can be considered analogously to the moment factorisation property of independent random variables. The normalized trace operator, which is equal to the first moment of the spectral density
\begin{equation}
\psi(H) = \frac{1}{N}\text{Tr}H = \frac{1}{N}\sum_{i=1}^{N}\lambda_{i} = \int_{\lambda \in \mathcal{D}} d\mu(\lambda)\lambda
\end{equation}
We say matrices $A \& B$ for which $\psi(A) = \psi(B) = 0$\footnote{We can always consider the transform $A - \psi(A)I$} are free if they satisfy for any integers $n_{1}..n_{k}$ with $k \in \mathbb{N}^{+}$
\begin{equation}
\psi(A^{n_{1}}B^{n_{2}}A^{n_{3}}B^{n_{4}}) = \psi(A^{n_{1}})\psi(B^{n_{2}})\psi(A^{n_{3}})\psi(A^{n_{4}})
\end{equation}

	\begin{theorem}
		\label{theorem:rmtbookwork}
		The extremal eigenvalues $[\lambda_{1}',\lambda_{P}']$ of the matrix sum $\mA+\mB/\sqrt{P}$, where $\mA \in \mathbb{R}^{P\times P}$ is a matrix of finite rank $r$ with extremal eigenvalues $[\lambda_{1},\lambda_{P}]$ and $\mB \in \mathbb{R}^{P\times P}$ is a GOE matrix with element variance $\sigma_{\epsilon}^{2}$ are given by
		\begin{equation}
			\lambda'_{1} = \left\{\begin{array}{lr}
				\lambda_{1} + \frac{\sigma_{\epsilon}^{2}}{\lambda_{1}}, & \text{if } \lambda_{1} > \sigma_{\epsilon}\\
				2 \sigma_{\epsilon}, & \text{otherwise } \\
			\end{array}\right\} 
, \thinspace
        \lambda'_{P} = \left\{\begin{array}{lr}
        				\lambda_{P} + \frac{\sigma_{\epsilon}^{2}}{\lambda_{n}}, & \text{if } \lambda_{n} < -\sigma_{\epsilon}\\
        				-2 \sigma_{\epsilon}, & \text{otherwise } \\
        			\end{array}\right\}
        		\end{equation}
	\end{theorem}

The Stijeles transform of Wigners semi circle law, can be written as \citep{tao2012topics}
\begin{equation}
    \mathcal{S}_{W}(z) = \frac{z \pm \sqrt{z^{2}-4\sigma^{2}}}{2\sigma^{2}}
\end{equation}
from the definition of the Blue transform, we hence have
\begin{equation}
\begin{aligned}
\label{eq:usingtheblue}
    & z = \frac{\mathcal{B}_{W}(z) \pm \sqrt{\mathcal{B}^{2}_{W}(z)-4\sigma^{2}}}{2\sigma^{2}}\\
    & (2\sigma^{2}z-\mathcal{B}_{W}(z))^{2} = \mathcal{B}^{2}_{W}(z)-4\sigma^{2} \\
    & \therefore \mathcal{B}_{W}(z) = \frac{1}{z} + \sigma^{2}z \\
    & \therefore \mathcal{R}_{W}(z) =  \sigma^{2}z\\
\end{aligned}
\end{equation}

Computing the $\mathcal{R}$ transform of the rank 1 matrix $H_{true}$, with largest non-trivial eigenvalue $\beta$, on the effect of the spectrum of a matrix $A$, using the Stieltjes transform we easily find following \citep{bun2017cleaning} that
\begin{equation}
    \mathcal{S}_{H_{true}}(u) = \frac{1}{N}\frac{1}{u-\beta}+\bigg(1-\frac{1}{N}\bigg)\frac{1}{u} = \frac{1}{u}\bigg[1 + \frac{1}{N}\frac{\beta}{1-u^{-1}\beta}\bigg] 
\end{equation}
We can use perturbation theory similar to in equation \eqref{eq:usingtheblue} to find the blue transform which to leading order gives
\begin{equation}
\begin{aligned}
    & \mathcal{B}_{H_{true}}(\omega) = \frac{1}{w} + \frac{\beta}{N(1-\omega \beta)} + \mathcal{O}(N^{-2}) \\
   &  \mathcal{R}_{H_{true}}(\omega) = \frac{\beta}{N(1-\omega \beta)} + \mathcal{O}(N^{-2})  \\
\end{aligned}
\end{equation}
setting $\omega = \mathcal{S}_{M}(z)$
\begin{equation}
    z = \mathcal{B}_{H_{true}}(\mathcal{S}_{M}(z)) + \frac{\beta}{N(1-\beta \mathcal{S}_{M}(z))} + \mathcal{O}(N^{-2})
\end{equation}
using the ansatz of $\mathcal{S}_{M}(z) = \mathcal{S}_{0}(z) + \frac{\mathcal{S}_{1}(z)}{N} + \mathcal{O}(N^{-2})$ we find that $\mathcal{S}_{0}(z) = \mathcal{S}_{\epsilon(w)}(z)$ and using that $\mathcal{B}'_{M}(z) = 1/g'(z)$ , we conclude that
\begin{equation}
    \mathcal{S}_{1}(z) = -\frac{\beta \mathcal{S'}_{\epsilon(w)}(z)}{1-\mathcal{S}_{\epsilon(w)}(z)\beta}
\end{equation}
and hence
\begin{equation}
    \mathcal{S}_{M}(z) \approx \mathcal{S}_{\epsilon(w)}(z) - \frac{1}{N}\frac{\beta \mathcal{S'}_{\epsilon(w)}(z)}{1-\mathcal{S}_{\epsilon(w)}(z)\beta}
\end{equation}
and hence in the large $N$ limit the correction only survives if $\mathcal{S}_{\epsilon(w)}(z) = 1/\beta$

\begin{equation}
\begin{aligned}
    & \mathcal{S}_{\epsilon(w)}(z) = \frac{1}{\beta} \\
    & \frac{2\sigma^{2}}{\beta} = z \pm \sqrt{z^{2}-4\sigma^{2}} \\
    & \therefore z = \beta + \frac{\sigma^{2}}{\beta}  \\
    \end{aligned}
\end{equation}
clearly for $\beta \rightarrow -\beta$ we have
\begin{equation}
    z = -\beta - \frac{\sigma^{2}}{\beta}
\end{equation}
\eqref{eq:spectralbroadeningwignernoise} follows simply from applying the correct normalisation constant in the definition of the Gaussian Orthogonal Ensemble which they use in \citet{arous1997large}.
\end{document}